\documentclass[runningheads]{llncs}
\usepackage{graphicx}
\usepackage{booktabs}
\usepackage{makecell}
\usepackage{amsmath}
\usepackage{enumitem}
\usepackage{svg}
\usepackage[utf8]{inputenc}
\usepackage[english]{babel}
\usepackage{tabularx}
\usepackage{hyperref}
\usepackage{bibentry}
\usepackage{multirow}
\usepackage{microtype}
\usepackage{footnote}
\usepackage{wrapfig}
\usepackage{float}

\usepackage{threeparttable}

\newcommand{\bestst}[1]{\textbf{#1}}
\newcommand{\bestnd}[1]{\underline{#1}}
\newcommand{\bestrd}[1]{\underline{\textit{#1}}}

\usepackage{array}
\newcolumntype{H}{>{\setbox0=\hbox\bgroup}c<{\egroup}@{}}

\begin{document}
\title{Point-Unet: A Context-aware Point-based Neural Network for Volumetric Segmentation}
\titlerunning{Point-Unet: A Point-based Neural Network for Volumetric Segmentation}

\author{
Ngoc-Vuong Ho\inst{1}\and
Tan Nguyen\inst{1} \and
Gia-Han Diep\inst{3} \and
Ngan Le\inst{4} \and
Binh-Son Hua\inst{1,2}
}

%index{Ho, Ngoc-Vuong}
%index{Nguyen, Tan}
%index{Diep, Gia-Han}
%index{Le, Ngan}
%index{Hua, Binh-Son}

\authorrunning{Ho et al.}

\institute{VinAI Research, Vietnam \\ \email{\{v.vuonghn,v.tannh10,v.sonhb\}@vinai.io}\and
VinUniversity \and
University of Science, VNU-HCM, Vietnam \\ \email{ han.diep@ict.jvn.edu.vn}\and
Department of Computer Science and Computer Engineering, \\ University of Arkansas, Fayetteville USA 72701\\
\email{thile@uark.edu} }
\maketitle              % typeset the header of the contribution
\begin{abstract} 
Medical image analysis using deep learning has recently been prevalent, showing great performance for various downstream tasks including medical image segmentation and its sibling, volumetric image segmentation. 
Particularly, a typical volumetric segmentation network strongly relies on a voxel grid representation which treats volumetric data as a stack of individual voxel `slices', which allows learning to segment a voxel grid to be as straightforward as extending existing image-based segmentation networks to the 3D domain. 
However, using a voxel grid representation requires a large memory footprint, expensive test-time and limiting the scalability of the solutions.
In this paper, we propose \emph{Point-Unet}, a novel method that incorporates the efficiency of deep learning with 3D point clouds into volumetric segmentation. 
Our key idea is to first predict the regions of interest in the volume by learning an attentional probability map, which is then used for sampling the volume into a sparse point cloud that is subsequently segmented using a point-based neural network. 
We have conducted the experiments on the medical volumetric segmentation task with both a small-scale dataset Pancreas and large-scale datasets BraTS18, BraTS19, and BraTS20 challenges. 
A comprehensive benchmark on different metrics has shown that our context-aware Point-Unet robustly outperforms the SOTA voxel-based networks at both accuracies, memory usage during training, and time consumption during testing. 
Our code is available at \url{https://github.com/VinAIResearch/Point-Unet}.

\end{abstract}

\section{Introduction}
\label{sec:intro}
% Context
Medical image segmentation has played an important role in medical analysis and is widely developed for many clinical applications. 
Although deep learning can achieve accuracy close to human performance for many computer vision tasks on 2D images, it is still challenging and limited for applying to medical imaging tasks such as volumetric segmentation. 
Existing voxel-based neural networks for volumetric segmentation have prohibitive memory requirements: nnNet \cite{nonewnet} uses a volume patch size of $160\times192\times128$, which requires a GPU with 32GB of memory for training to achieve the state-of-the-art performance~\cite{Partially_unet}. 
To mitigate high memory usage, some previous work resort to workarounds such as using smaller grid size (e.g., $25^3$ and $19^3$ in DeepMedic \cite{deepmedic}) for computation, resulting in degraded performance.  

\begin{figure}[t!]
\centering
\includegraphics[width=\linewidth]{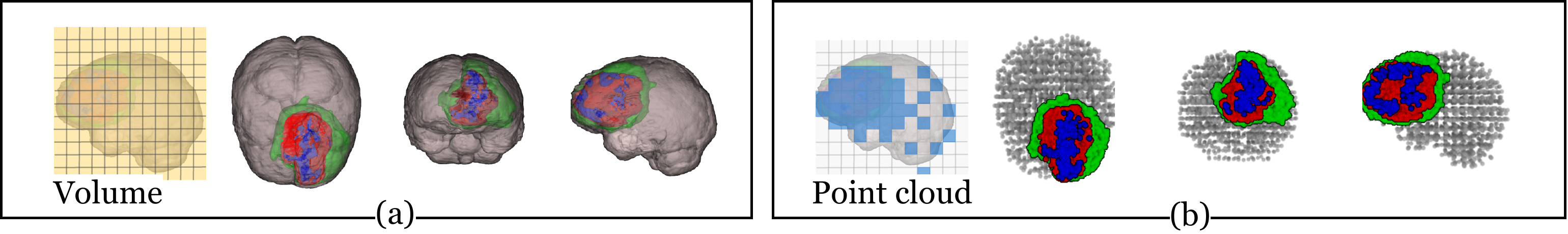}
\caption{(a): a 2D voxel grid and a segmentation rendered by volume rendering; (b): a PC from the grid and the point-based segmentation results.}
\label{fig:compare}
\end{figure}

In this work, we propose to leverage the 3D PC representation for the problem of medical volumetric segmentation as inspired by recent success in 3D point cloud (PC) analysis~\cite{qi2017pointnet,qi2017pointnet++,li2018pointcnn,wang2018edgeconv,hu2020randla}.
Having a PC representation is advantageous in that we can have fine-grained control of the segmentation quality, i.e., to sample the volume and focus points at the important areas. 
PCs are also suitable for capturing global features that are challenging and costly to have with a regular voxel grid. 
A summary of the difference between PC and voxel grid on an MRI image is shown in Figure~\ref{fig:compare}.

Our so-called \emph{Point-Unet} is a point-based volumetric segmentation framework with three main modules: the saliency attention, the context-aware sampling, and the point-based segmentation module. The saliency attention module takes a volume as input and predicts an attentional probability map that guides the context-aware point sampling in the subsequent module to transform the volume into a PC. The point-based segmentation module then processes the PCs and outputs the segmentation, which is finally fused back to the volume to obtain the final segmentation results.

% highligh contribution
In summary, our main contributions in this work are: (1) Point-Unet, a new perspective and formulation to solve medical volumetric segmentation using a PC representation; 
(2) A saliency proposal network to extract an attentional probability map which emphasizes the regions of interests in a volume;
(3) An efficient context-aware point sampling mechanism for capturing better local dependencies within regions of interest while maintaining global relations; 
(4) A comprehensive benchmark that demonstrates the advantage of our point-based method over other SOTA voxel-based 3D networks at both accuracies, memory usage during training, and inference time.

\section{Related Work}
\noindent\textbf{Volumetric Segmentation.}
Deep learning-based techniques, especially CNNs, have shown excellent performance in the volumetric medical segmentation.
Early methods include the standard Unet~\cite{3Dunet}, Vnet~\cite{VNet}, and then DeepMedic~\cite{deepmedic}, which improves robustness with multi-scale segmentation. 
Recently, by utilizing hard negative mining to achieve the final voxel-level classification, \cite{Kao_brats2018} improved the patch-based CNNs performance. 
KD-Net~\cite{KD-Net_Brats2018} fused information from different modalities through knowledge distillation. 
Instead of picking the best model architecture, \cite{Ensemble} ensembled multiple models which were trained on different datasets or different hyper-parameters. By extending U-Net with leaky ReLU activations and instance normalization, nnNet~\cite{nonewnet} obtained the second-best performance on BraTS18. aeUnet \cite{brats_autoencoder}, the top-performing method in BraTS18, employed an additional branch to reconstruct the input MRI on top of a traditional encoder-decoder 3D CNN architecture. 
The top performance of BraTS19 is \cite{2stage_brats2019}, which is a two-stage cascaded U-Net. The first stage had a U-Net architecture. In the second stage, the output of the first stage was concatenated to the original input and fed to a similar encoder-decoder to obtain the final segmentation.

\noindent\textbf{Point Cloud Segmentation.} 
In 3D deep learning, the semantic segmentation task can be solved by directly analyzing PCs data.
%, which can be easily acquired from LiDAR systems, SAR systems and RGB-D cameras. 
Many point-based techniques have been recently developed for PC semantic segmentation~\cite{qi2017pointnet,superpoint_2018,hu2020randla}. 
%PointNet~\cite{qi2017pointnet} was a pioneering deep learning framework which performed directly on point. 
PointNet~\cite{qi2017pointnet} used MLPs to learn the representation of each point, whereas the global representation was extracted by applying a symmetric function like max pooling on the per-point features. 
%However, no local structure information within neighboring points was learnt by PointNet \cite{qi2017pointnet}. 
PointNet++~\cite{qi2017pointnet++} was then developed to address the lack of local features by using a hierarchy of PointNet itself to capture local geometric features in a local point neighborhood. %Dynamic Graph CNNs \cite{wang2018edgeconv} generalized the idea of PointNet++ by further considering edges in the neighborhood, forming a dynamic graph at each point neighborhood when extracting local features. 
PointCNN \cite{li2018pointcnn} used a $\mathcal{X}$-transformation to learn features from unstructured PCs. 
In order to extract richer edge features \cite{superpoint_2018} proposed SuperPoint Graph where each superpoint was embedded in a PointNet and then refined by RNNs. 
Inspired by the idea of the attention mechanism, \cite{GAC_2019} proposed a graph-based convolution with attention to capture the structural features of PCs while avoiding feature contamination between objects. 
Recently, RandLA-Net \cite{hu2020randla} has achieved SOTA performance on semantic segmentation of large point clouds by leveraging random sampling at inference. 
In this work, we aim to bring the efficiency of deep learning with point clouds into volumetric segmentation for medical 3D data. 

\section{Proposed Point-Unet}
\begin{figure}[t!]
\centering
\includegraphics[width=\textwidth]{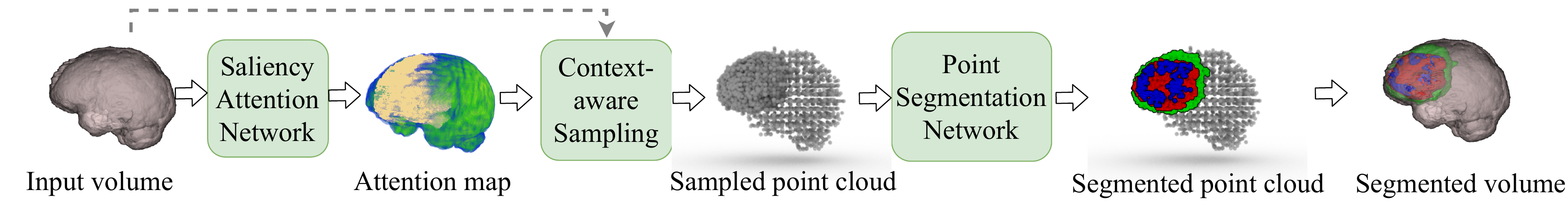}
\caption{Point-Unet takes a volume as input and consists of 3 modules: saliency attention network, context-aware sampling and point segmenation network.}
\label{fig:flowchart}
\end{figure}

Our proposed Point-Unet for volumetric segmentation contains three modules i.e.,to saliency attention module, context-aware sampling and point-based segmentation module. The overall architecture is given in Fig.~\ref{fig:flowchart}.

\begin{figure*}[!t]
\centering
\includegraphics[width=\textwidth]{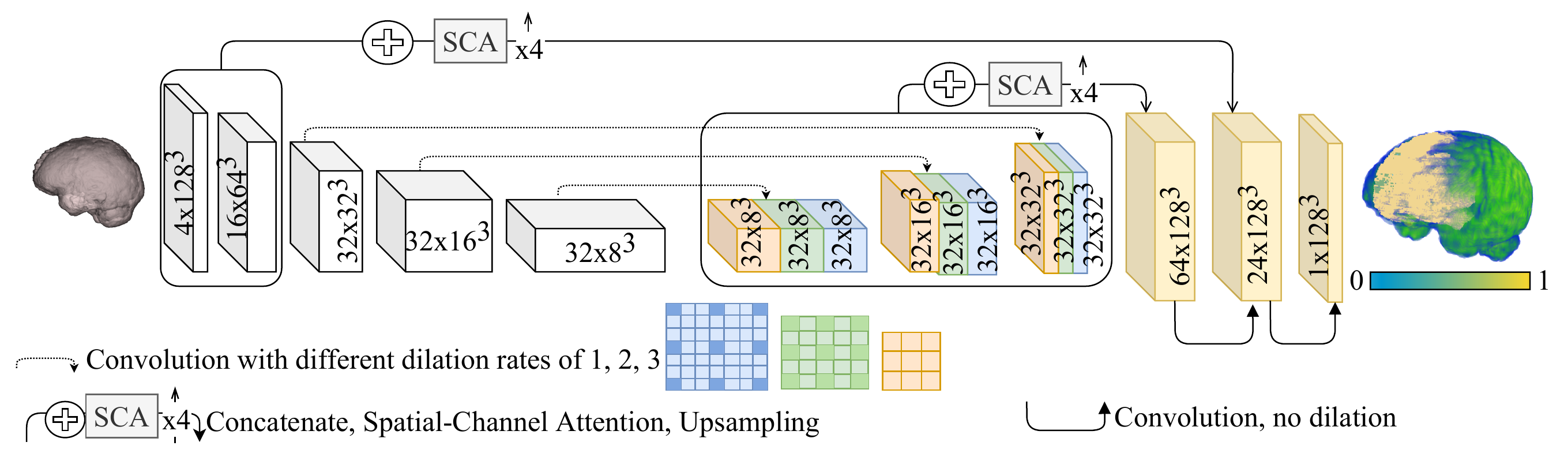}
\caption{Our proposed saliency attention network.}
\label{fig:SA_flowchart}
\end{figure*}

\subsection{Saliency Attention}
\label{subsec:saliency}

Our saliency attention network is leveraged by \cite{sca_gate,attention_model}, and designed as contextual pyramid to capture multi-scale with multi-receptive-field at high-level features. The network is illustrated in Fig.~\ref{fig:SA_flowchart} and contains two high-level layers and two low-level layers. At the high-level features, we adopt atrous convolution with different dilation rates set to 1, 2, and 3 to capture multi receptive field context information. The feature maps from different atrous layers are then combined by concatenation while the smaller ones are upsampled to the largest one. Then, we combine them by cross channel concatenation and channel-wise attention (SCA) \cite{sca_gate} as the output of the high-level feature extraction. At the low-level features, we apply SCA \cite{sca_gate} to combine two low-level features maps after upsampling the smaller ones. The high-level feature is then upsampled and combined with the low-level feature to form a feature map at original resolution. 

\subsection{Context-aware Sampling}
\label{subsec:sampling}
Random sampling (RS) used in the original RandLA-Net~\cite{hu2020randla} has been successfully applied into 3D shapes, but it is not a good sampling technique for medical volumetric data because of following reasons: (i) there is no mechanism in RS to handle intra-imbalance; (ii) topological structure is important in medical analysis but there is no attention mechanism in RS to focus on the object boundary which are very weak in medical images; (iii) RS samples points all over the data space, it may skip small objects while objects of interest in medical are relatively small; (iv) volumetric data is large and RS requires running inference multiple times which is time consuming. 
\begin{figure*}[!t]
\centering
\includegraphics[width=\textwidth]{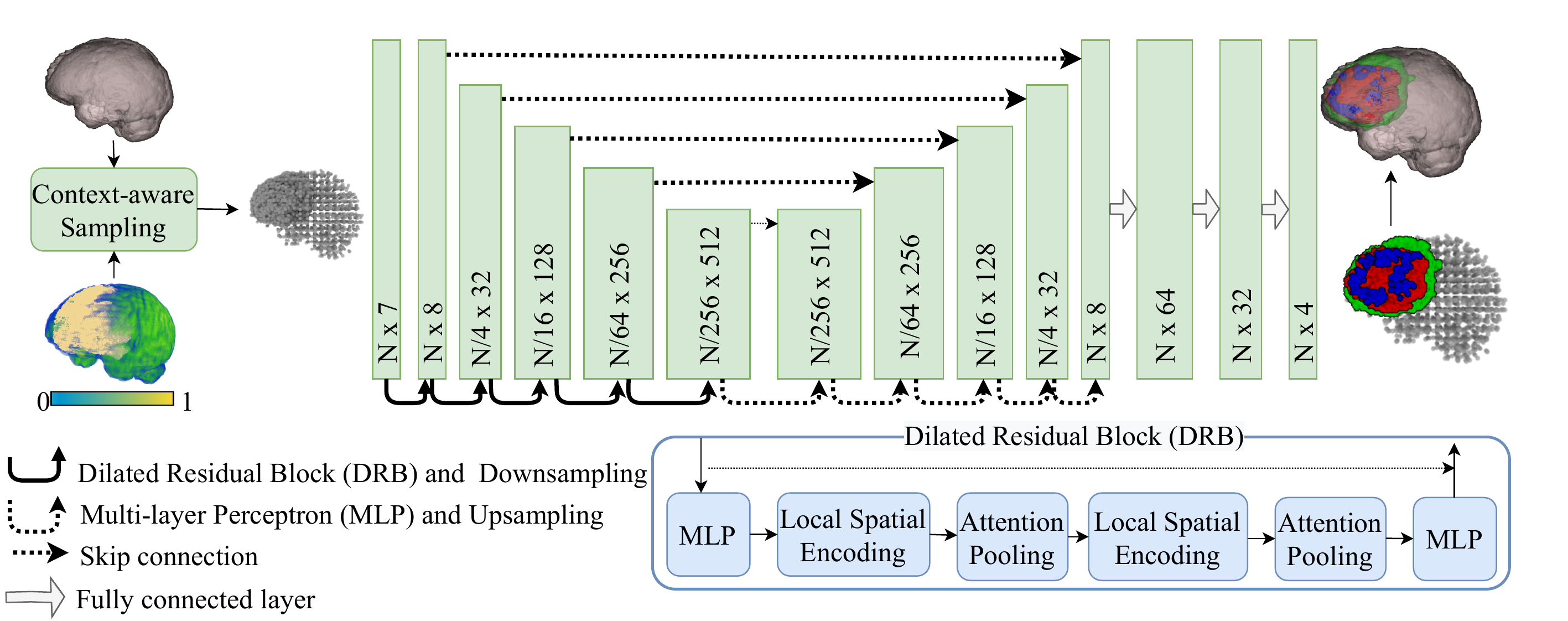}
\caption{Our proposed Point-Unet segmentation on volumetric data.}
\label{fig:Point_volume_flowchart}
\end{figure*}

Our context-aware sampling is designed to take all such limitations into account. Here the main conceptual differences are that our sampling is a single pass and our sampling only samples i.e., it is necessary just to sample the volume and perform the inference once. 
Our context-aware sampling is as follows. Firstly, we sample the points identified by the attentional probability map obtained by the saliency attention module. On the salient region where the probabilities are higher, we densely sample points to better learn contextual local representation. In the non-salient region, we apply random sampling to keep global relations. By doing that, our context-aware sampling can capture better local dependencies within regions of interest while maintaining global relations.

\subsection{Point-based Segmentation}
\label{subsec:point-based segmentaiton}
Given a volumetric data, we first sample a PC by Subsection~\ref{subsec:sampling}.%using the attentional probability map from the previous step so that the PC is a sparse representation of the original volumetric data. 
Our point-based segmentation departs from the original design of RandLA-Net~\cite{hu2020randla} in that we introduce a context-aware sampling technique for effectively sampling PCs from a volume. We also redesign RandLA-Net under Unet architecture~\cite{3Dunet} together with an appropriate loss to better fit the task of medical volumetric segmentation. The architecture of the proposed point-based segmentation network is illustrated in Fig.~\ref{fig:Point_volume_flowchart}. Our network takes $N$ points as its input $\{p_i\}_{i=1}^N$, each point $p_i = \{x^i, f^i\}$ where $x^i$ is a tuple of coordinates in sagittal, coronal, transverse planes and $f^i$ is a tuple of point features including point intensity in $T_1$, $T_2$, $T_{ce}$, and $Flair$ modalities. 

The input PC is first processed through encoder path, which contains sequences Dilated Residual Block (DRB) and downsampling. Each DRB includes multiple units of multi-layer perceptrons (MLP), local spatial encoding and attentive pooling stacks and the DRBs are connected through skip-connections as proposed in RandLA-Net~\cite{hu2020randla}. The output of the encoder path is then processed through the decoder path, which consists of a sequence of MLP and upsampling. The network also makes use of the skip connection while upsampling to learn feature representation at multi-scale. Finally, the decoder output is passed through three fully connected layers with drop-out regularization. 

Far apart from RandLA-Net\cite{hu2020randla} and most existing segmentation networks, which use the cross-entropy (CE) loss, we utilize Generalized Dice Loss (GDL) \cite{soft_dice} for training.
GDL \cite{soft_dice} has been proven to be efficient at dealing with data imbalance problems that often occur in medical image segmentation. GDL is computed as
$
    GDL = 1 - 2\frac{\sum_{l=1}{w_l\sum_n{r_{ln}p_{ln}}}}{\sum_{l=1}{w_l\sum_n{(r_{ln}+p_{ln})}}}
$
 where $w_l =  \frac{1}{\sum_n^N{r_{ln}}}$ is used to provide invariance to different label set properties. $R$ is gold standard with value at voxel $n^{th}$ denoted as $r_n$. $P$ the predicted segmentation map with value at voxel $n^{th}$ denoted as $p_n$. For class $l$, the groundtruth and predicted labels are $r_{ln}$ and $p_{ln}$.

%Discuss difference between our sampling with cascade network

\section{Experimental Results}

We evaluate our method and compare to the state-of-the-art methods on two datasets: Pancreas and BraTS. 
Pancreas~\cite{pancreas} contains 82 abdominal contrast enhanced 3D CT scans. The CT scans have resolutions of 512x512 pixels with varying pixel sizes and slice thickness between 1.5--2.5 mm.
BraTS~\cite{Brats} consists of a large-scale brain tumor dataset. The training set includes 285/335/369 patients and validation set contains 66/125/125 patients in BraTS18/BraTS19/BraTS20. 
Each image is registered to a common space, sampled to an isotropic with skull-stripped and has a dimension of $240 \times 240 \times 155$. 
For point sampling in our training and inference, we sample 180,000 points in Pancreas, and 350,000 points in BraTS. 
%The ground truth has been manually labeled by experts. In the experimental results, we have conducted the benchmarks on both training set and validation set. 
%As for local evaluation, we randomly partition the training set into 2 subsets: 295 subjects for training and 74 subjects for testing. As for online evaluation, we train Point-Unet on 369 subjects and test on 125 cases from online validation set. 

% ==================2018=====================

% \footnotetext[1]{Reproduce the results on the network trained with 100 epochs.}
% \footnotetext[2]{The results reported in the paper.}
% \footnotetext[3]{We choose the model with as similar batch size as ours.}
% \footnotetext[4]{We choose single model with 190 epoches, stage 1. The best model at Brats2020}
% \footnotetext[5]{The first place of BraTS19. We choose the Ensemble of 5-fold}
% \footnotetext[6]{The second place on BraTS19}

\begin{savenotes}
\begin{table}[!t]
\setlength\tabcolsep{0.5pt}
%Row: The $1^{st}$ group: voxel-based methods, $2^{nd}$ group: point-based methods. Column: The $1^{st}$ part: training set, the $2^{nd}$ part: validation set.
\caption{Comparison on BraTS18. The \textbf{best}, \underline{second best} and \textit{\underline{third best}} are highlighted.}
\label{tab:brats18}
\centering
\resizebox{1\textwidth}{!}{
\begin{tabular}{c |cc|Hc||c|cc|Hc@{}}
\toprule
\textbf{BraTS18} &\multicolumn{4}{c||}{\textbf{Offline validation set}}& \multicolumn{5}{c}{\textbf{Online validation set}} \\ 
\toprule
Methods  &  \multicolumn{2}{c|}{Dice score $\uparrow$ }& \multicolumn{2}{c||}{HD95 $\downarrow$ }& Methods  &  \multicolumn{2}{c|}{Dice score $\uparrow$ }& \multicolumn{2}{c}{HD95 $\downarrow$ } \\ 
 & {ET/WT/TC} & {AVG} & {ET/WT/TC} & {AVG}& & {ET/WT/TC} & {AVG} & {ET/WT/TC} & {AVG} \\ \hline
 3DUnet\cite{baseline}  &66.82/81.19/77.58 &75.20 &6.84/12.50/8.61   &9.32 &3DUnet\cite{Unet_baseline}    &72.05/84.24/76.41   & 77.56     &14.02/23.36/21.97   & 17.83    \\ [1ex]  
 
3DUNet\cite{3DUnet_Brats2018}  & 68.43/\bestrd{89.91}/\underline{86.77} &81.70 & 5.33/\underline{5.56}/\textbf{5.11} &\bestst{5.33} &KD-Net\cite{KD-Net_Brats2018} &   71.67/81.45/76.98  &76.70 &-----/-----/----- &----- \\ [1ex]

h-Dense\cite{h-Dense_Brats2018}  & 70.69/89.51/82.76 &80.99 &  6.24/6.04/6.95 &6.41 &DenseNet\cite{DenseUnet} &\textit{\underline{80.00}}/90.00/82.00 &84.00 &------/------/------ &----- \\ [1ex]

DMF\cite{DMFNet_Brats2018}  & \bestrd{76.35}/89.09/82.70 &\bestnd{82.71} & -----/-----/----- & ----- & aeUnet\cite{brats_autoencoder}\footnotemark[2] & \textbf{81.45}/\textit{\underline{90.42}}/\textbf{85.96}   & \textbf{85.94}    &   \underline{3.81}/\textbf{4.48}/8.28  & \underline{5.52}  \\ [1ex]
aeUnet\cite{brats_autoencoder} &75.31/85.69/81.98 & 80.99 &6.68/11.45/7.80  &8.64  & aeUnet\cite{brats_autoencoder}\footnotemark[1] & 72.60/85.02/77.33  &  78.32 & 13.62/23.42/18.69    &18.58  \\ [1ex]

S3D\cite{S3D-UNet_brats2018}  & 73.95/88.81/\bestrd{84.42} &\bestrd{82.39} &\underline{4.63}/5.89/5.67 &\bestnd{5.40} & S3D \cite{S3D-UNet_brats2018}      &  74.93/89.35/83.09  &82.56 & \textit{\underline{4.43}}/\textit{\underline{4.72}}/\underline{7.75} & \textit{\underline{5.63}} \\ [1ex]

nnNet\cite{nonewnet} &\underline{76.65}/81.57/84.21 &80.81  &8.46/18.02/11.29  &12.59 &nnNet\cite{nonewnet}\footnotemark[2]  & 79.59/\textbf{90.80}/\underline{84.32} &\textit{\underline{84.90}} &\textbf{3.12}/4.79/8.16 &\textbf{5.36}  \\ [1ex]

KaoNet\cite{Kao_brats2018}  & 73.50/\underline{90.20}/81.30 &81.67 & 5.43/\textbf{5.40}/6.93  &\bestrd{5.92} &nnNet\cite{nonewnet}\footnotemark[1]  &75.01/82.23/81.84 &79.69 &5.47/8.21/\underline{7.75} & 7.14 \\ [1ex]

\midrule
RandLA\cite{hu2020randla} &70.05/88.13/80.32  &79.50  &8.11/5.68/\underline{5.30}  &6.36   & RandLA\cite{hu2020randla}& 73.05/87.30/76.94 & 79.10 &5.44/\underline{4.57}/\textbf{7.36} & 5.79 \\ [1ex]

\textbf{Ours}&\textbf{80.76}/\textbf{90.55}/\textbf{87.09} &\textbf{86.13} &\textbf{2.96 }/8.40/6.69 &6.01 & \textbf{Ours} & \underline{80.97}/\underline{90.50}/\textit{\underline{84.11}} & \underline{85.19} &4.30/6.63/\textit{\underline{7.97}} & 	6.30   \\

%\midrule
%3D U-Net-adadelta + Mixup     & 0.7477& 0.8735 & 0.7875 & 33.5278 & 9.5473   & 21.8207                 \\
% + postprocessed     & 0.7674& 0.8734 & 0.7874 & 30.3873 & 9.5682   & 21.8346                \\
\hline
\end{tabular}
}
\end{table}
\end{savenotes}

%==========================2019===================
% \footnotetext[1]{Reproduce the results on the network trained with 100 epochs.}
% \footnotetext[2]{The results reported in the paper.}
% \footnotetext[3]{We choose the model with as similar batch size as ours.}
\begin{savenotes}
\begin{table}[ht]
\setlength\tabcolsep{0.5pt}
\caption{Comparison on BraTS19. The \textbf{best}, \underline{second best} and \textit{\underline{third best}} are highlighted.}
\label{tab:brats19}
\centering
\resizebox{1\textwidth}{!}{
\begin{tabular}{c|cc|Hc||c|cc|Hc@{}}
\toprule
\textbf{BraTS19} & \multicolumn{4}{c||}{\textbf{Offline validation set}}& \multicolumn{5}{c}{\textbf{Online validation set}} \\ 
\toprule
Methods  &  \multicolumn{2}{c|}{Dice score $\uparrow$ }& \multicolumn{2}{c||}{HD95 $\downarrow$ }& Methods  &  \multicolumn{2}{c|}{Dice score $\uparrow$ }& \multicolumn{2}{c}{HD95 $\downarrow$ } \\ 
 & {ET/WT/TC} & {AVG} & {ET/WT/TC} & {AVG}& & {ET/WT/TC} & {AVG} & {ET/WT/TC} & {AVG} \\ \midrule
 3DUnet\cite{baseline}  &67.74/80.17/78.92  &75.61  &9.81/17.59/7.68    &11.69 &3DUnet\cite{Unet_baseline} &64.26/79.65/72.07 &72.00 &17.27/30.09/27.48     & 23.68     \\ [1ex]

nnNet\cite{nonewnet}  &79.46/81.13/\bestrd{87.08} &82.67 &4.70/9.79/5.76 &6.75 &nnNet\cite{nonewnet}\footnotemark[1] &70.42/81.53/78.22 &76.72 &6.47/9.21/9.23  & 8.30     \\ [1ex]

aeUnet\cite{brats_autoencoder} &80.55/86.26/85.78 &84.19 &7.14/16.85/8.84  &10.94 &aeUnet \cite{brats_autoencoder}\footnotemark[1] & 64.81/83.02/74.48 & 74.10&  17.59/25.91/19.79  & 21.75     \\ [1ex]

HNF\cite{HR-Net_Brats2019}   & \bestrd{80.96}/\bestrd{91.12}/86.40 &\bestrd{86.16} &-----/-----/----- &----- &HNF\cite{HR-Net_Brats2019} & \bestst{81.16}/\bestst{91.12}/\bestnd{84.52}    & \bestst{85.60} &3.49/4.13/5.25 & \bestnd{3.81}\\ [1ex]
%SynNet~\cite{Synthetic_Brats2019} & 2020 & 79.26 & 91.65 & 90.76 & 3.50 & 5.70 & 3.40 \\

N3D\cite{Unet_Brats2019}  & \underline{83.0}/\textbf{91.60}/\underline{88.80} &\bestnd{87.35} &\underline{3.07}/\textbf{4.01}/\textbf{3.67} &\bestst{3.58} &Synth\cite{Synthetic_Brats2019} &76.65/\bestrd{89.65}/79.01 &81.77 &4.6/6.9/8.4  &5.75\\ [1ex]

& & & & &\shortstack{2stage~\cite{2stage_brats2019}}\footnotemark[5] &\bestrd{79.67}/\bestnd{90.80}/\bestst{85.89} &\bestst{85.45} &3.12/4.35/5.69 &\bestst{3.74}\\ [1ex]

& & & & &3Unet\cite{Unet_Brats2019} & 73.70/89.40/80.70 & 81.27 & 5.99/5.68/7.36 & 5.84\\ [1ex]

& & & & &3DSe~\cite{3DSemanticSeg_Brats2019} & \bestnd{80.00}/89.40/\bestrd{83.40} & \bestrd{84.27} & 3.92/5.89/6.56 & \bestrd{4.91}\\ [1ex]

& & & & &\shortstack{Bag-trick\cite{BagTrick_Brats2019}}\footnotemark[6] & 70.20/88.30/80.00  & 79.50 & 4.77/5.08/6.47 & 4.93\\ \midrule

RandLA\cite{hu2020randla}  &76.68/89.01/84.81 &83.50 &3.36/5.45/\underline{4.54} &\bestnd{4.45} &RandLA\cite{hu2020randla} &70.77/86.95/74.27 &70.77 &6.84/5.84/8.60 & 7.09 \\ [1ex]

\textbf{Ours}   &\textbf{85.67}/\underline{91.18}/\textbf{90.10}  &\bestst{88.98} &\textbf{2.90}/\underline{6.92}/4.95 &\bestrd{4.92} &\textbf{Ours} &79.01/87.63/79.70 &82.11 &6.93/12.97/11.27 &10.39  \\

%\midrule
%3D U-Net-adadelta + Mixup     & 0.7477& 0.8735 & 0.7875 & 33.5278 & 9.5473   & 21.8207                 \\
% + postprocessed     & 0.7674& 0.8734 & 0.7874 & 30.3873 & 9.5682   & 21.8346                \\
\bottomrule
\end{tabular}
}
\end{table}
\end{savenotes}

%===================2020==========================================

\begin{savenotes}
\begin{table}[!ht]
\setlength\tabcolsep{0.5pt}
\caption{Comparison on BraTS20. The \textbf{best}, \underline{second best} and \textit{\underline{third best}} are highlighted.}
\label{tab:brats20}
\centering
\resizebox{1\textwidth}{!}{
\begin{tabular}{c |cc|Hc||c|cc|Hc@{}}
\toprule
\textbf{BraTS20}& \multicolumn{4}{c||}{\textbf{Offline validation set}}& \multicolumn{5}{c}{\textbf{Online validation set}} \\ 
\toprule
Methods  &  \multicolumn{2}{c|}{Dice score $\uparrow$ }& \multicolumn{2}{c||}{HD95 $\downarrow$ }& Methods  &  \multicolumn{2}{c|}{Dice score $\uparrow$ }& \multicolumn{2}{c}{HD95 $\downarrow$ } \\ 
 & {ET/WT/TC} & {AVG} & {ET/WT/TC} & {AVG}& & {ET/WT/TC} & {AVG} & {ET/WT/TC} & {AVG} \\ \midrule
 
3DUnet\cite{baseline}    & 66.92/82.86/72.98 &74.25 & 44.96/17.37/28.24 &30.19 & 3DUNet \cite{Unet_baseline}   &67.66/87.35/79.30 & 78.10 &47.10/7.90/8.47  & 21.16\\ [1ex]

nnNet\cite{nonewnet}  &\underline{73.64}/80.99/\underline{81.60}  &\bestnd{78.74} & 21.97/12.32/8.71 &\bestrd{14.33} & nnNet \cite{nonewnet}\footnotemark[1] & 68.69/81.34/78.06  &  78.03 & 48.01/9.45/15.43  & 24.30\\ [1ex]

aeUnet\cite{brats_autoencoder}
& \bestrd{71.31}/\bestrd{84.72}/\bestrd{79.02}  &\bestrd{78.35} &19.64/14.5/12.15 &15.43 &nnUNet\cite{nnunet_Brats2020} \footnotemark[3] &\textit{\underline{77.67}}/\textbf{90.60}/\textbf{84.26} &\textbf{84.18} &35.10/\textbf{4.89}/\textbf{5.91} & 15.30\\ [1ex]

& & & & & aeNet\cite{brats_autoencoder}\footnotemark[1]  &64.00/83.16/74.66  &73.95 &52.71/26.77/22.27 &33.91\\ [1ex]

& & & & & Cascade\cite{cascade_Brats2020}\footnotemark[4]& \underline{78.81}/\underline{89.92}/\textit{\underline{82.06}} & \bestrd{83.60} &\underline{23.71}/\underline{5.65}/\underline{6.66} & \textit{\underline{12.00}}\\ [1ex]

& & & & &KiUNet\cite{KiU-Net_Brats2020} &73.21/87.60/73.92  &78.24 &\textbf{6.32}/8.94/9.89 & \textbf{8.38 }\\ [1ex] \midrule

RandLA\cite{hu2020randla} & 67.40/\underline{87.74}/76.85 & 77.33 &\underline{9.66}/\textbf{5.25}/\textbf{6.17} &\bestst{7.03} &RandLA\cite{hu2020randla}& 66.31/88.01/77.03 &77.17 & 35.38/\textbf{4.89}/9.68  & 16.65 \\ [1ex]

\textbf{Ours} & \textbf{76.43}/\textbf{89.67}/\textbf{82.97} &\bestst{83.02} &\textbf{ 6.12}/\underline{10.19}/\underline{8.48} &\bestnd{8.26} &\textbf{Ours} &\textbf{78.98}/\textit{\underline{89.71}}/\underline{82.75} &\underline{83.81} &\textit{\underline{21.43}}/\textit{\underline{6.73}}/\textit{\underline{7.05}}  & \underline{11.73}\\
\bottomrule
\end{tabular}
}
\end{table}
\end{savenotes}

\footnotetext[1]{Reproduce the results on the network trained with 100 epochs.}
\footnotetext[2]{The results reported in the paper.}
\footnotetext[3]{We choose the model with as similar batch size as ours.}
\footnotetext[4]{We choose single model with 190 epoches, stage 1. The best model at Brats2020}
\footnotetext[5]{The first place of BraTS19. We choose the Ensemble of 5-fold}
\footnotetext[6]{The second place on BraTS19}

\noindent
\textbf{Evaluation Setup:} For each dataset, we experiment on both offline validation set and online validation set. The evaluation on the offline validation set is conducted locally by partitioning the training set into training subset (80\%) and evaluation subset (20\%). We train the network on TensorFlow empowered by Tensorpack \cite{huf2015tensorpack} to speed up the training. We use Momentum optimizer with momentum value $0.9$, learning rate $0.01$ with decay, and batch size $2$. The model is trained on an NVIDIA Tesla V100 32GB GPU for 100 epochs.
For BraTS, we compare our method with SOTA voxel-based results, and with RandLA-Net~\cite{hu2020randla}, the SOTA point-based segmentation method. 
Note that our evaluation is done on volume as our point-based segmentation results can be transferred directly to the volume without further processing thanks to our sampling scheme.

\noindent
\textbf{Evaluation Results: }
The evaluation on Brats is given in Tables~\ref{tab:brats18},~\ref{tab:brats19},~\ref{tab:brats20} for BraTS 2018, BraTS 2019, and BraTS 2020, respectively. Whereas the evaluation on Pancreas is given in Table~\ref{tab:pancreas}. On Brats, there are two groups of methods corresponding to SOTA voxel-based and point-based reported in each table. While our Point-Unet achives either better or competitive performance compared to SOTA voxel-based methods, it is better than SOTA point-based method (RandLA-Net) at both Dice score and HD95. For nnNet~\cite{nonewnet} and aeNet~\cite{brats_autoencoder}, we train and reproduce the resutls. It shows that the reproduced results are always lower than the ones reported which were postprocessed. Without postprocessing, our results outperforms both nnNet~\cite{nonewnet} and aeNet~\cite{brats_autoencoder}. In other words, our Point-Unet obtains SOTA performance on both offline validation set and online validation set without postprocessing. Not only on the large-scale dataset such as Brats, our Point-Unet also obtains the SOTA performance on small-scale dataset such as Pancreas as shown Table~\ref{tab:pancreas}. 

Please also refer to the supplementary material for full comparisons on Pancreas, an ablation study of our network, and other implementation details.

%===========Pancreas===========
% \begin{table}[!t]
% \caption{Comparison on Pancreas dataset.}
% \label{tab:pancreas}
% \setlength{\tabcolsep}{0.75em}
% \centering
% \begin{tabular}{l | lll | l}
% \toprule
% & Zhou \cite{zhou2017fixed} & Yu \cite{yu2018recurrent}  & Oktay \cite{oktay2018attention}& \textbf{Ours} \\ 
% \midrule
% Avg. Dice  & 82.37 $\pm$ 5.68 & 84.50 $\pm$ 4.97 & 83.1 $\pm$ 3.8 & \textbf{85.68 }$\pm$5.96\\ 
% %Min    & 62.43 & 60.0 & - & 59.81\\ 
% %Max    & 90.85  & 90.1 & - & 94.47\\ 
% %\# Folds    & 4 & 4 & 1 & 4\\ 
% \bottomrule
% \end{tabular}
% \end{table}

\begin{table}[!t]
\caption{Dice score comparison on Pancreas dataset.}
\label{tab:pancreas}
\setlength{\tabcolsep}{0.75em}
\centering
\begin{tabular}{|l|l||l|l|}
\toprule
 Method & Average $\uparrow$& Method &  Average $\uparrow$\\ \midrule
Oktay et al.~\cite{oktay2018attention}  &  83.10 $\pm$ 3.80 & Yu et al.~\cite{yu2018recurrent} &  84.50 $\pm$ 4.97 \\
\midrule
Zhu et al~\cite{zhu20183d} & 84.59 $\pm$ 4.86 & 
\textbf{Ours} & \textbf{85.68 }$\pm$5.96 \\ \bottomrule
\end{tabular}
\end{table}

\begin{figure}[!t]
\begin{minipage}{0.5\textwidth}
    \includegraphics[width=\linewidth]{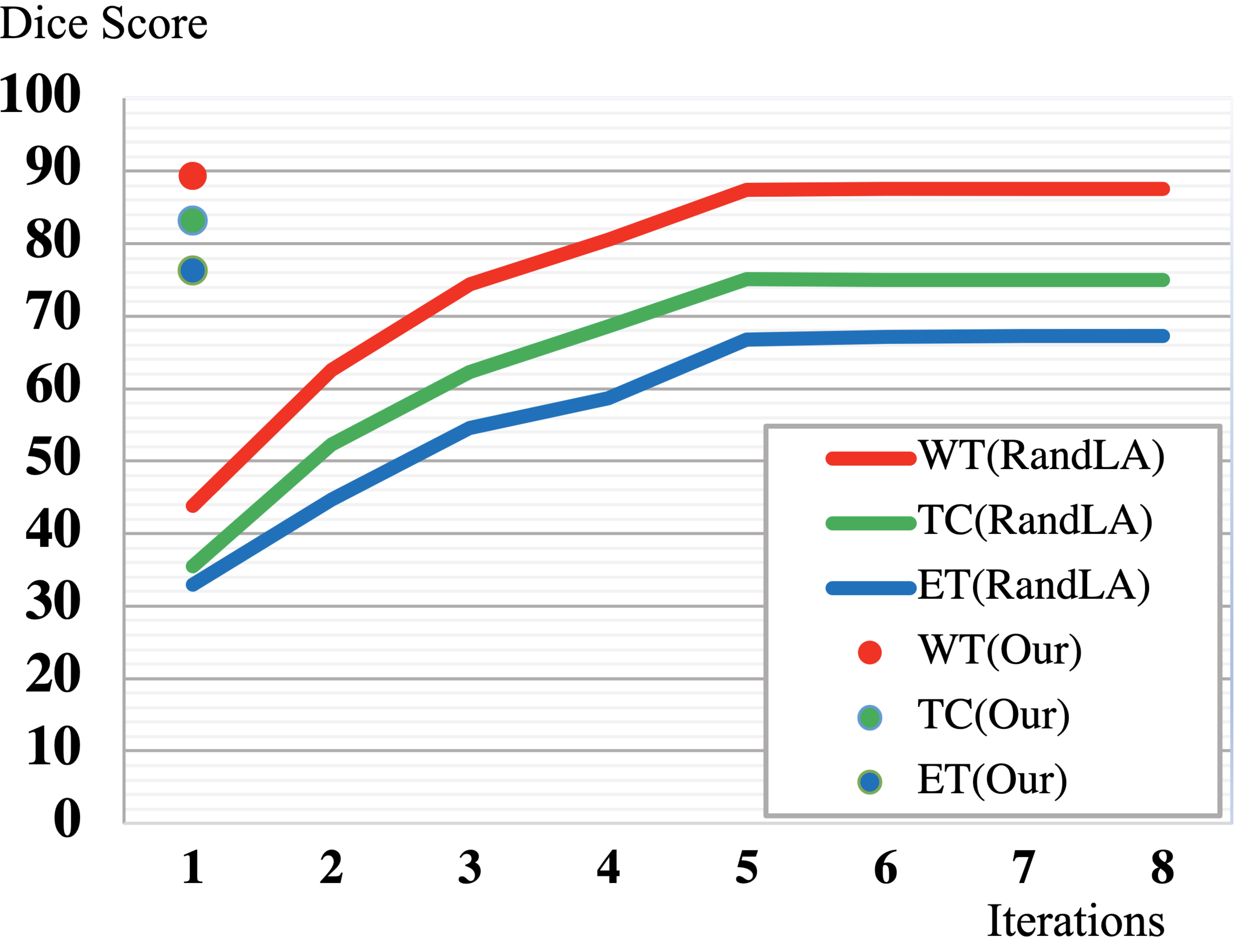}
    \centerline{(a)}
\end{minipage} 
\begin{minipage}{0.5\textwidth}
    \resizebox{\textwidth}{!}{
    \begin{tabular}{Hccrr}
    \toprule
    &Networks           & Patch size  &  \multicolumn{1}{c}{\begin{tabular}[c]{@{}c@{}}Memory $\downarrow$\end{tabular}}  &   \multicolumn{1}{c}{\begin{tabular}[c]{@{}c@{}}Inference $\downarrow$\end{tabular}} \\ \midrule
    \multirow{9}{*}{\rotatebox{90}{volume-based}} & \multirow{3}{*}{\shortstack{3DUnet\\baseline\\~\cite{baseline}}}   & $128\times128\times128$ & 8.75 GB & 7.80 s    \\
                              & & $160\times192\times128$ & 16.70 GB  & 0.28 s     \\
                              & & $240\times240\times144$ & 32.00 GB  & 0.23 s \\ [1ex]
    & \multirow{3}{*}{\shortstack{nnNet\\ \cite{nonewnet}}} & $128\times128\times128$ &     7.20 GB  &  55.30 s \\ 
                              & & $160\times192\times128$ &  11.10 GB  &  26.50 s  \\
                             &  & $240\times240\times144$ &  21.70 GB  &  2.30 s \\ [1ex]
    & \multirow{3}{*}{\shortstack{aeUnet \\ \cite{brats_autoencoder}}} & $128\times128\times128$ &    17.21 GB  & 110.40 s  \\
                              & & $160\times192\times128$ &    31.42 GB &  78.80 s \\
                              & & $240\times240\times144$ &    $>$48.00 GB  & 7.10 s  \\ \midrule
    % \multirow{3}{*}{Saliency Network} & $128\times128\times128$ &    17.22GB  &  24.8 \\
    %                           & $160\times192\times128$ &    31.44GB & 0.33  \\
    %                           & $240\times240\times144$ &    xGB  & 0.24  \\\midrule             
    \multirow{2}{*}{\rotatebox{90}{\shortstack{point-\\based}}} & \shortstack{RandLA\cite{hu2020randla}}                  & $240\times240\times155$ & 15.98 GB &  8.00 s  \\ [1ex]
    & \textbf{(Ours)}              & $240\times240\times155$ & 17.22 GB & 1.24 s  \\ \bottomrule
    \end{tabular}
    }
    \centerline{(b)}
\end{minipage}

\caption{Performance analysis. (a) With a single inference, our Point-Unet outperforms RandLA-Net, which requires multiple iterations at inference. (b) Memory requirement for training with batch size 1 and inference time with difference volume patch sizes.}
\label{fig:performance_analysis}
\end{figure}

\noindent
\textbf{Performance Analysis:} 
Fig.~\ref{fig:performance_analysis}(a) shows the comparison between our Point-Unet against RandLA-Net~\cite{hu2020randla} in terms of the number of iterations performed during inference. The experiment is conducted on BraTS20 offline validation set. The performance of RandLA-Net with RS strategy highly depend on the number of iterations. It reaches the best performance when RS covers the entire volume, which requires up to eight iterations. By using context-aware sampling, our PC covers regions of interest in just a single iteration while outperforming RandLA-Net. 
Figure~\ref{fig:performance_analysis}(b) provides the memory requirement during training with batch size set to 1 on different input volume patch sizes. 
We also measure inference time by three runs and then take the averages. 
In general, for voxel-based networks, a smaller patch size requires less memory during training but it takes more time at inference. By contrast, point-based networks including our Point-Unet and RandLA-Net~\cite{hu2020randla} require much less memory to handle the entire volume while keeping the inference time plausible.

\section{Conclusion}
In this work, we introduced Point-Unet, a point-based framework for volumetric segmentation. We tested our framework on the problem of brain tumor and pancreas segmentation and showed that our point-based neural segmentation is robust, scalable, and more accurate than existing voxel-based segmentation methods. 
%Our work is not without limitations. In theory, our method can work for any volumetric data, but in practice, it is the most effective for data that the segmentation is centered at some regions of interests (salient regions) while remaining regions do not need to be labeled. 
Future investigations might aim for better techniques for volume-point sampling and label reconstruction. 
Techniques for segmentation boundary adaptive sampling~\cite{kirillov2020pointrend} and attention-based convolution~\cite{woo2018cbam} are also potential extensions for performance improvement. 

\vspace{3mm}
\noindent
\textbf{Acknowledgment:} 
This material is based upon work supported by the National Science Foundation under Award No. OIA-1946391.

\vspace{3mm}
\noindent
\textbf{Disclaimer:}
Any opinions, findings, and conclusions or recommendations expressed in this material are those of the author(s) and do not necessarily reflect the views of the National Science Foundation.

%\newpage
\bibliographystyle{splncs04}
\bibliography{paper2053}

\begin{thebibliography}{10}
\providecommand{\url}[1]{\texttt{#1}}
\providecommand{\urlprefix}{URL }
\providecommand{\doi}[1]{https://doi.org/#1}

\bibitem{Partially_unet}
Brügger, R., Baumgartner, C.F., Konukoglu, E.: A partially reversible u-net for memory-efficient volumetric image segmentation. MICCAI  (2019)

\bibitem{DMFNet_Brats2018}
Chen, C., Liu, X., Ding, M., Zheng, J., Li, J.: 3d dilated multi-fiber network for real-time brain tumor segmentation in mri. In: MICCAI (2019)

\bibitem{sca_gate}
{Chen}, L., {Zhang}, H., {Xiao}, J., {Nie}, L., {Shao}, J., {Liu}, W., {Chua}, T.: Sca-cnn: Spatial and channel-wise attention in convolutional networks for image captioning. In: CVPR (2017)

\bibitem{S3D-UNet_brats2018}
Chen, W., Liu, B., Peng, S., Sun, J., Qiao, X.: S3d-unet: Separable 3d u-net for brain tumor segmentation. In: MICCAI Brainlesion (2019)

\bibitem{Ensemble}
Feng, X., Tustison, N.J., Patel, S.H., Meyer, C.H.: Brain tumor segmentation using an ensemble of 3d u-nets and overall survival prediction using radiomic features. Frontiers in computational neuroscience  \textbf{14}(25) (2020)

\bibitem{Synthetic_Brats2019}
Hamghalam, M., Lei, B., Wang, T.: Brain tumor synthetic segmentation in 3d multimodal mri scans. In: MICCAI Brainlesion (2020)

\bibitem{KD-Net_Brats2018}
Hu, M., Maillard, M., Zhang, Y., Ciceri, T., La~Barbera, G., Bloch, I., Gori, P.: Knowledge distillation from multi-modal to mono-modal segmentation networks. In: MICCAI (2020)

\bibitem{hu2020randla}
Hu, Q., Yang, B., Xie, L., Rosa, S., Guo, Y., Wang, Z., Trigoni, N., Markham, A.: Randla-net: Efficient semantic segmentation of large-scale point clouds. In: CVPR (2020)

\bibitem{huf2015tensorpack}
Huf, P., Carminati, J.: Tensorpack: a maple-based software package for the manipulation of algebraic expressions of tensors in general relativity. In: J Phys Conf Ser. vol.~633 (2015)

\bibitem{nnunet_Brats2020}
Isensee, F., Jaeger, P.F., Full, P.M., Vollmuth, P., Maier-Hein, K.H.: nnu-net for brain tumor segmentation (2020)

\bibitem{3DUnet_Brats2018}
Isensee, F., Kickingereder, P., Wick, W., Bendszus, M., Maier-Hein, K.H.: Brain tumor segmentation and radiomics survival prediction: Contribution to the brats 2017 challenge. In: MICCAI Brainlesion (2018)

\bibitem{nonewnet}
Isensee, F., Kickingereder, P., Wick, W., Bendszus, M., Maier-Hein, K.H., van Walsum, T.: No new-net. In: MICCAI Brainlesion (2019)

\bibitem{HR-Net_Brats2019}
Jia, H., Xia, Y., Cai, W., Huang, H.: Learning high-resolution and efficient non-local features for brain glioma segmentation in mr images. In: MICCAI (2020)

\bibitem{2stage_brats2019}
Jiang, Z., Ding, C., Liu, M., Tao, D.: Two-stage cascaded u-net: 1st place solution to brats challenge 2019 segmentation task. In: MICCAI Brainlesion (2020)

\bibitem{baseline}
Kamnitsas, K., Bai, W., Ferrante, E., McDonagh, S., Sinclair, M., Pawlowski, N., Rajchl, M., Lee, M., Kainz, B., Rueckert, D., et~al.: Ensembles of multiple models and architectures for robust brain tumour segmentation. In: MICCAI Brainlesion. Springer (2017)

\bibitem{Kao_brats2018}
Kao, P.Y., Ngo, T., Zhang, A., Chen, J.W., Manjunath, B.S.: Brain tumor segmentation and tractographic feature extraction from structural mr images for overall survival prediction. In: MICCAI Brainlesion (2019)

\bibitem{kirillov2020pointrend}
Kirillov, A., Wu, Y., He, K., Girshick, R.: Pointrend: Image segmentation as rendering. In: CVPR (2020)

\bibitem{superpoint_2018}
{Landrieu}, L., {Simonovsky}, M.: Large-scale point cloud semantic segmentation with superpoint graphs. In: CVPR (2018)

\bibitem{h-Dense_Brats2018}
{Li}, X., {Chen}, H., {Qi}, X., {Dou}, Q., {Fu}, C.W., {Heng}, P.A.: H-denseunet: Hybrid densely connected unet for liver and tumor segmentation from ct volumes. IEEE TMI  \textbf{37}(12) (2018)

\bibitem{li2018pointcnn}
Li, Y., Bu, R., Sun, M., Chen, B.: Pointcnn: Convolution on x-transformed points. NIPS  (2018)

\bibitem{cascade_Brats2020}
Lyu, C., Shu, H.: A two-stage cascade model with variational autoencoders and attention gates for mri brain tumor segmentation (2020)

\bibitem{deepmedic}
McKinley, R., Meier, R., Wiest, R.: Ensembles of densely-connected cnns with label-uncertainty for brain tumor segmentation. In: MICCAI Brainlesion (2019)

\bibitem{Brats}
Menze, B.H., Jakab, A., Bauer, et~al.: The multimodal brain tumor image segmentation benchmark (brats). IEEE TMI  \textbf{34}(10) (2015)

\bibitem{VNet}
Milletari, F., Navab, N., Ahmadi, S.: V-net: Fully convolutional neural networks for volumetric medical image segmentation. CoRR  (2016)

\bibitem{brats_autoencoder}
Myronenko, A.: 3d mri brain tumor segmentation using autoencoder regularization. In: MICCAI Brainlesion (2019)

\bibitem{3DSemanticSeg_Brats2019}
Myronenko, A., Hatamizadeh, A.: Robust semantic segmentation of brain tumor regions from 3d mris. In: MICCAI Brainlesion (2020)

\bibitem{oktay2018attention}
Oktay, O., Schlemper, J., Folgoc, L.L., et~al.: Attention u-net: Learning where to look for the pancreas. arXiv preprint arXiv:1804.03999  (2018)

\bibitem{qi2017pointnet}
Qi, C.R., Su, H., Mo, K., Guibas, L.J.: Pointnet: Deep learning on point sets for 3d classification and segmentation. In: CVPR (2017)

\bibitem{qi2017pointnet++}
Qi, C.R., Yi, L., Su, H., Guibas, L.J.: Pointnet++: Deep hierarchical feature learning on point sets in a metric space. In: NIPS (2017)

\bibitem{3Dunet}
{Ronneberger}, O., {Fischer}, P., {Brox}, T.: {U-Net: Convolutional Networks for Biomedical Image Segmentation}. arXiv e-prints p. arXiv:1505.04597 (May 2015)

\bibitem{pancreas}
Roth, H.R., Farag, A., et~al.: Data from pancreas-ct (2016)

\bibitem{soft_dice}
Sudre, C.H., Li, W., Vercauteren, T., Ourselin, S., Jorge~Cardoso, M.: Generalised dice overlap as a deep learning loss function for highly unbalanced segmentations. In: DLMIA/ML-CDS@MICCAI (2017)

\bibitem{KiU-Net_Brats2020}
Valanarasu, J., Sindagi, V.A., Hacihaliloglu, I., Patel, V.M.: Kiu-net: Overcomplete convolutional architectures for biomedical image and volumetric segmentation (2020)

\bibitem{Unet_Brats2019}
Wang, F., Jiang, R., Zheng, L., Meng, C., Biswal, B.: 3d u-net based brain tumor segmentation and survival days prediction. In: MICCAI Brainlesion (2020)

\bibitem{Unet_baseline}
Wang, G., Li, W., Ourselin, S., Vercauteren, T.: Automatic brain tumor segmentation using cascaded anisotropic convolutional neural networks. In: MICCAI Brainlesion (2018)

\bibitem{GAC_2019}
{Wang}, L., {Huang}, Y., {Hou}, Y., {Zhang}, S., {Shan}, J.: Graph attention convolution for point cloud semantic segmentation. In: CVPR (2019)

\bibitem{wang2018edgeconv}
Wang, Y., Sun, Y., Liu, Z., Sarma, S.E., Bronstein, M.M., Solomon, J.M.: Dynamic graph cnn for learning on point clouds. ACM Transactions on Graphics  (2019)

\bibitem{woo2018cbam}
Woo, S., Park, J., Lee, J.Y., Kweon, I.S.: Cbam: Convolutional block attention module. In: ECCV (2018)

\bibitem{DenseUnet}
Yogananda, B., et~al., C.G.: A fully automated deep learning network for brain tumor segmentation. Tomography  \textbf{6}(2) (2020)

\bibitem{yu2018recurrent}
Yu, Q., Xie, L., Wang, Y., Zhou, Y., Fishman, E.K., Yuille, A.L.: Recurrent saliency transformation network: Incorporating multi-stage visual cues for small organ segmentation. In: CVPR (2018)

\bibitem{attention_model}
{Zhao}, T., {Wu}, X.: Pyramid feature attention network for saliency detection. In: CVPR (2019)

\bibitem{BagTrick_Brats2019}
Zhao, Y.X., Zhang, Y.M., Liu, C.L.: Bag of tricks for 3d mri brain tumor segmentation. In: MICCAI Brainlesion (2019)

\bibitem{zhu20183d}
Zhu, Z., Xia, Y., Shen, W., Fishman, E., Yuille, A.: A 3d coarse-to-fine framework for volumetric medical image segmentation. In: 3DV. pp. 682--690. IEEE (2018)

\end{thebibliography}


\begin{thebibliography}{10}
\providecommand{\url}[1]{\texttt{#1}}
\providecommand{\urlprefix}{URL }
\providecommand{\doi}[1]{https://doi.org/#1}

\bibitem{cai2017improving}
Cai, J., Lu, L., Xie, Y., Xing, F., Yang, L.: Improving deep pancreas segmentation in ct and mri images via recurrent neural contextual learning and direct loss function. arXiv preprint arXiv:1707.04912  (2017)

\bibitem{dou20163d}
Dou, Q., Chen, H., Jin, Y., Yu, L., Qin, J., Heng, P.A.: 3d deeply supervised network for automatic liver segmentation from ct volumes. In: MICCAI. pp. 149--157. Springer (2016)

\bibitem{hu2020randla}
Hu, Q., Yang, B., Xie, L., Rosa, S., Guo, Y., Wang, Z., Trigoni, N., Markham, A.: Randla-net: Efficient semantic segmentation of large-scale point clouds. In: CVPR (2020)

\bibitem{nonewnet}
Isensee, F., Kickingereder, P., Wick, W., Bendszus, M., Maier-Hein, K.H., van Walsum, T.: No new-net. In: MICCAI Brainlesion (2019)

\bibitem{baseline}
Kamnitsas, K., Bai, W., Ferrante, E., McDonagh, S., Sinclair, M., Pawlowski, N., Rajchl, M., Lee, M., Kainz, B., Rueckert, D., et~al.: Ensembles of multiple models and architectures for robust brain tumour segmentation. In: MICCAI Brainlesion. Springer (2017)

\bibitem{mrbrains18}
Kuijf, H.J., Bennink, E.: Grand challenge on mr brain segmentation at miccai 2018 (2018), \url{http://mrbrains18.isi.uu.nl/}

\bibitem{brats_autoencoder}
Myronenko, A.: 3d mri brain tumor segmentation using autoencoder regularization. In: MICCAI Brainlesion (2019)

\bibitem{oktay2018attention}
Oktay, O., Schlemper, J., Folgoc, L.L., et~al.: Attention u-net: Learning where to look for the pancreas. arXiv preprint arXiv:1804.03999  (2018)

\bibitem{roth2015deeporgan}
Roth, H.R., Lu, L., Farag, A., Shin, H.C., Liu, J., Turkbey, E.B., Summers, R.M.: Deeporgan: Multi-level deep convolutional networks for automated pancreas segmentation. In: MICCAI. pp. 556--564. Springer (2015)

\bibitem{roth2016spatial}
Roth, H.R., Lu, L., Farag, A., Sohn, A., Summers, R.M.: Spatial aggregation of holistically-nested networks for automated pancreas segmentation. In: MICCAI. pp. 451--459. Springer (2016)

\bibitem{roth2018spatial}
Roth, H.R., Lu, L., Lay, N., Harrison, A.P., Farag, A., Sohn, A., Summers, R.M.: Spatial aggregation of holistically-nested convolutional neural networks for automated pancreas localization and segmentation. Medical image analysis  \textbf{45},  94--107 (2018)

\bibitem{yu2018recurrent}
Yu, Q., Xie, L., Wang, Y., Zhou, Y., Fishman, E.K., Yuille, A.L.: Recurrent saliency transformation network: Incorporating multi-stage visual cues for small organ segmentation. In: CVPR (2018)

\bibitem{zhang2016coarse}
Zhang, Y., Ying, M.T., Yang, L., Ahuja, A.T., Chen, D.Z.: Coarse-to-fine stacked fully convolutional nets for lymph node segmentation in ultrasound images. In: International Conference on Bioinformatics and Biomedicine (BIBM). pp. 443--448. IEEE (2016)

\bibitem{attention_model}
{Zhao}, T., {Wu}, X.: Pyramid feature attention network for saliency detection. In: CVPR (2019)

\bibitem{zhou2017fixed}
Zhou, Y., Xie, L., Shen, W., Wang, Y., Fishman, E.K., Yuille, A.L.: A fixed-point model for pancreas segmentation in abdominal ct scans. In: MICCAI (2017)

\bibitem{zhu20183d}
Zhu, Z., Xia, Y., Shen, W., Fishman, E., Yuille, A.: A 3d coarse-to-fine framework for volumetric medical image segmentation. In: 3DV. pp. 682--690. IEEE (2018)

\end{thebibliography}

\end{document}

\title{Supplementary Material for\\ Point-Unet: A Context-aware Point-based Neural Network for Volumetric Segmentation}
\titlerunning{Supplementary Material for Point-Unet}

\author{
Ngoc-Vuong Ho\inst{1}\and
Tan Nguyen\inst{1} \and
Gia-Han Diep\inst{3} \and
Ngan Le\inst{4} \and
Binh-Son Hua\inst{1,2}
}
%index{Ho, Ngoc-Vuong}
%index{Nguyen, Tan}
%index{Diep, Gia-Han}
%index{Le, Ngan}
%index{Hua, Binh-Son}
\authorrunning{Ho et al.}

\institute{VinAI Research, Vietnam \\ \email{\{v.vuonghn,v.tannh10,v.sonhb\}@vinai.io}\and
VinUniversity \and
University of Science, VNU-HCM, Vietnam \\ \email{ han.diep@ict.jvn.edu.vn}\and
Department of Computer Science at University of Arkansas in Fayetteville \\
\email{thile@uark.edu} }

\maketitle              % typeset the header of the contribution

% \begin{abstract} 
% In this supplemental document, we provide additional experiments to compare the random sampling and our context-aware sampling scheme. Particularly, we show that random sampling results in boundary artifacts and missing small objects that affect the final segmentation quality. By contrast, our context-aware sampling does not have these artifacts while it is able to capture small objects in most of the cases. We have conducted the experiments on the medical volumetric segmentation task with both a small-scale dataset Pancreas and large-scale datasets BraTS18, BraTS19, and BraTS20 challenges
% \keywords{Volumetric Segmentation \and Point Cloud}
% \end{abstract}

\section{Pancreas Results} 
We provide full comparisons on Pancreas dataset in Table~\ref{tab:pancreas_full}, which extends the results reported in the main paper.

\begin{table}[!h]
\caption{Dice score comparison on Pancreas dataset.}
\label{tab:pancreas_full}
\setlength{\tabcolsep}{0.75em}
\centering
\begin{tabular}{|l|l||l|l|}
\toprule
 Method & Average $\uparrow$& Method &  Average $\uparrow$\\ \midrule
Zhou et al. \cite{zhou2017fixed}  &  82.37 $\pm$ 5.68  & Roth et al.~\cite{roth2016spatial} & 78.01 $\pm$ 8.20  \\ 
Roth et al.~\cite{roth2015deeporgan} & 71.42 $\pm$ 10.11  & Roth et al.~ \cite{roth2018spatial} & 81.27 $\pm$ 6.27 \\ 
Oktay et al.~\cite{oktay2018attention}  &  83.10 $\pm$ 3.80  & Zhang et al.~\cite{zhang2016coarse} & 77.89 $\pm$ 8.52\\ 
Cai et al.~\cite{cai2017improving} &  82.40 $\pm$ 6.70    & Zhou et al.~\cite{zhou2017fixed} & 82.37 $\pm$ 5.68  \\
Zhu et al~\cite{zhu20183d} & 84.59 $\pm$ 4.86 & Dou et al.~\cite{dou20163d} & 82.25 $\pm$ 5.91 \\
\midrule
\textbf{Ours} & \textbf{85.68 }$\pm$5.96 & Yu et al.~\cite{yu2018recurrent} &  84.50 $\pm$ 4.97 \\ \bottomrule
\end{tabular}
\end{table}

\section{Ablation study}
We conducted an ablation study on the offline validation set of BraTS20. We provided variants of our model to demonstrate the significance of the volumetric saliency attention network, point-based segmentation network, and the GDL loss. The results are shown in Table~\ref{tab:ablation}, which demonstrates the importance of our proposed saliency network and point-based segmentation. 

\setlength{\tabcolsep}{11pt}
\renewcommand{\arraystretch}{1.25}
\begin{table}[h!]
\centering
\caption{Ablation study. We show dice measurements on variants of our method on the BraTS 2020 offline validation set.}
\label{tab:ablation}
\begin{tabular}{@{}l|c|c|c|c @{}}
\hline
\multirow{2}{*}{Method}   & \multicolumn{4}{c}{Dice score $\uparrow$} \\ \cline{2-5} 
&{ET} & {WT} & {TC} & {AVG} \\ \hline
E: 3D U-Net~\cite{baseline} & 66.92 & 82.86 & 72.98 & 74.25 \\ \hline
D: Attention 3D U-Net~\cite{attention_model} &  69.87 & 89.68 & 79.28 & 79.61  \\ \hline
C: Ours without GDL Loss &69.33 & 89.36 & 69.51 & 76.06  \\\hline
B: Ours without saliency (RandLANet)     & 67.40 & 87.74 & 76.85 & 77.33                 \\\hline
A: Ours (Point-Unet) &76.43 & 89.67 & 82.97 & 83.02 \\\hline
\end{tabular}
\end{table}

Particularly, model A is our reported model in Table 3 in the main paper. 
By removing the saliency attention network and only use point-based segmentation, we obtain model B that resembles RandLANet, the accuracy of which drops significantly (6\%). Model B was reported in Table 3 (paper) as RandLANet~\cite{hu2020randla}.
To verify the effectiveness of the generalized dice loss (GDL), we remove GDL and use cross entropy in the point-based network (model C), which also results in accuracy loss by almost 7\%. 
Compared to the traditional U-Net (3D U-Net~\cite{baseline}, model E) and an attention-based U-Net which we extended from~\cite{attention_model} (model D) that employs only volumetric segmentation, which is equivalent to our method without the point-based segmentation, our method also outperforms significantly. 
This verifies that both components (volumetric saliency attention network and point-based segmentation network) together with GDL contribute significantly in our method. 

\section{Details on Point Sampling}
We provide additional details for the point sampling step in our method.
In general, our proposed method has two main steps: saliency attention prediction and point-based segmentation. In our training, we train the saliency network (voxel segmentation) and point segmentation separately. We first train the saliency network with the target labels as binary: foreground that are union of all ground truth tumor regions, and background for remaining voxels. Dice loss is used for training the saliency network. After training the saliency network, we perform point sampling based on thresholding the confidence output from the saliency network to establish point clouds for segmentation.

\setlength{\tabcolsep}{11pt}
\renewcommand{\arraystretch}{1.25}
\begin{table}[h!]
\centering
\caption{Evaluation with different confidence thresholds for our point sampling scheme. The results are on BraTS 2020 offline set.}
\label{tab:threshold}
\begin{tabular}{l|c|c|c|c }
\hline
\multirow{2}{*}{Threshold}   & \multicolumn{4}{c}{Dice score $\uparrow$} \\ \cline{2-5} 
&{ET} & {WT} & {TC} & {AVG} \\ \hline
0.5  & 75.82 & 85.53 & 82.05 & 81.14  \\ \hline
0.6  & 75.90 & 86.43 & 82.26 & 81.53  \\ \hline
0.7  & 75.91 & 87.45 & 82.69 & 82.02                 \\\hline
0.8  & 76.04 & 88.45 & 82.88 & 82.46  \\\hline
\textbf{0.9}  & \textbf{76.43} & \textbf{89.67} & 82.97 & \textbf{83.02}  \\\hline
0.95  & 76.26 & 89.39 & 83.12 & 82.93 \\\hline
0.975  & 76.36 & 89.35 & 82.57 & 82.76 \\\hline
\end{tabular}
\end{table}

To generate the point cloud, we threshold the confidence output of the attention network (threshold 0.9), and voxels passing the threshold become foreground (FG) points. We randomly sample remaining voxels to obtain background (BG) points. The union of FG and BG points form an input point cloud for segmentation. Note that the FG already contains tumor regions, and the BG only provides additional context data for learning. The segmentation results of the FG can be simply used as the final tumor segmentation results, and *no resampling* from point cloud to volume is required. We also tested with different thresholds and found that it is quite insensitive to the model performance (~1-percent difference when varying the threshold in [0.6, 0.95]) as shown in Table~\ref{tab:threshold}.

\subsection{Random Sampling vs. Our Sampling}

\begin{figure}[!h]
\centering
\includegraphics[width=0.9\textwidth]{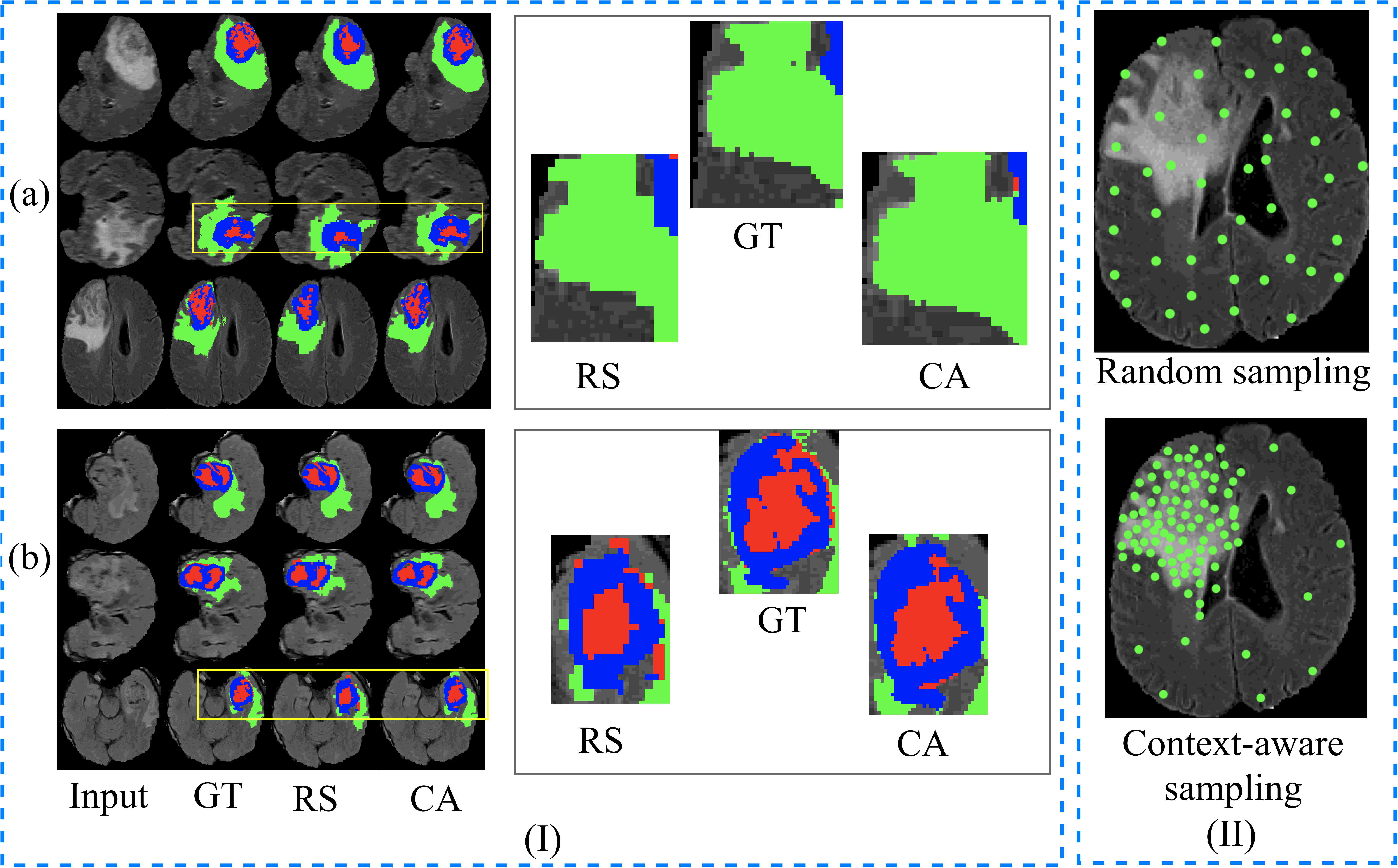}
\caption{Illustration on the \emph{small objects}. \textbf{RS}:  random sampling, \textbf{CA}: context-aware, \textbf{GT}: groundtruth. (a) and (b) are two different brain subjects. The enlarged view of the lesion is given on the right.}
\label{fig:RS_vis}
\end{figure}

The comparison between random sampling and our proposed context-aware sampling is given in Figure~\ref{fig:RS_vis}(II). 
Random sampling treats every pixel in the same manner, thus there is no mechanism to pay attention to boundaries as well as small objects to address the above limitations. As given in Figure~\ref{fig:RS_vis}(I), random sampling produces the results in zigzag artifact at boundaries while the object surface plays an important role in medical analysis which aims to understand the topologial strucutre. Furthermore, random sampling scheme may not always capture every sample in a population. This causes the missing of small objects during sampling and segmenting as given in Figure~\ref{fig:RS_vis}(I). Not only unable to address the aformentioned difficulties, random sampling also has some other restrictions. The limitations of random sampling is summarized as follows:
\begin{itemize}[leftmargin=*]
    \item Results in zigzag artifact at boundaries;
    \item Unable to capture small objects;
    \item Inference is costly: In order cover entire brain space, it requires performing random sampling serval times;
    \item Cannot guarantee the tumor area will be completely covered after many iterations at inference.
\end{itemize}
By contrast, our context-aware sampling addresses such issues by placing more samples at the region of interests while maintaining samples in the background regions.

\section{Visualization Result on BraTS}
\begin{figure}[h!]
\centering
\includegraphics[width=0.9\textwidth]{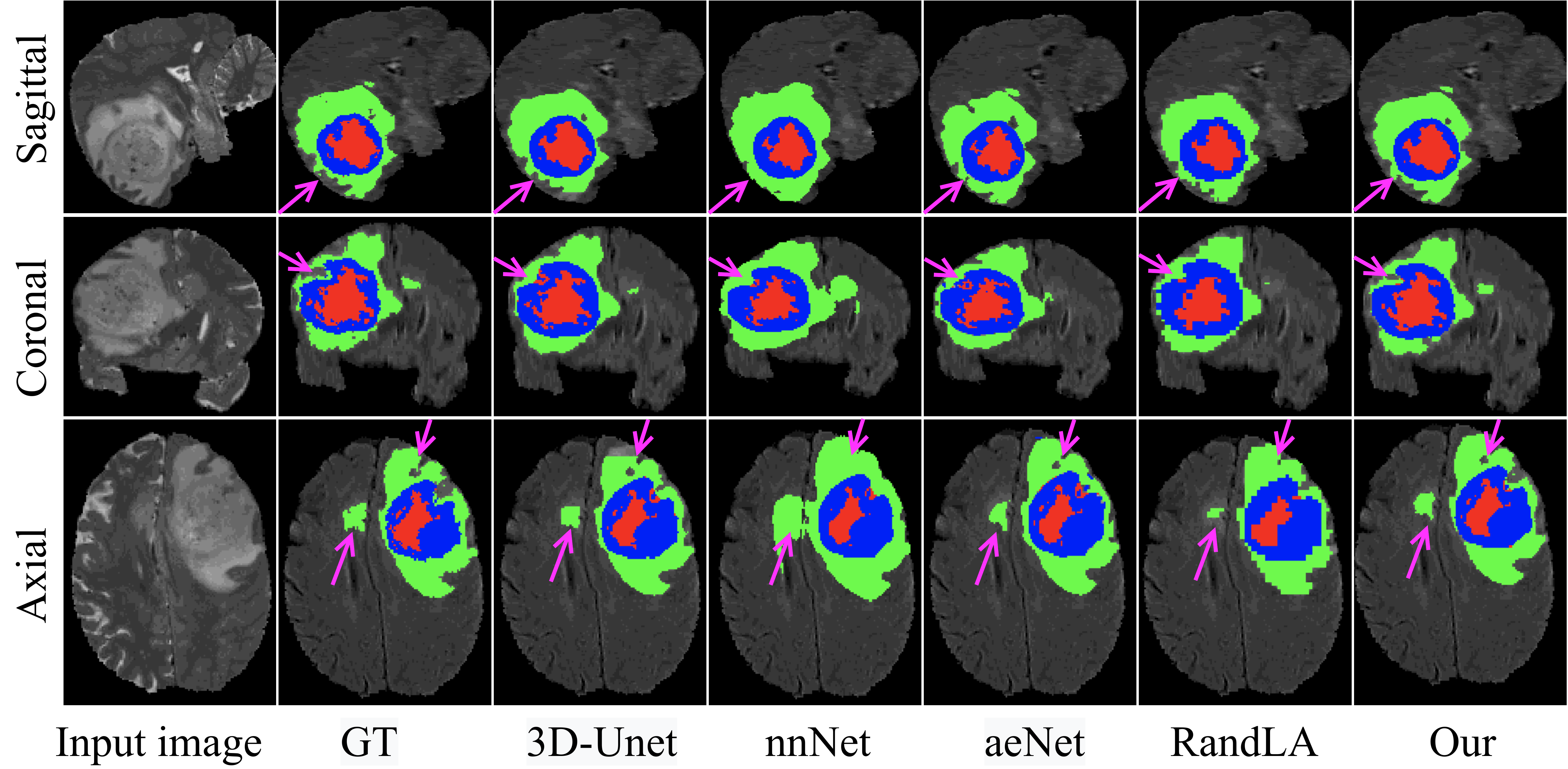}
\caption{Segmentation results on different planes ($1^{st}$ row: Sagittal, $2^{nd}$ row: Coronal, $3^{rd}$: Axial) generated by (a) source image (Flair modality); (b) groundtruth; (c) baseline 3DUnet\cite{baseline}; (d) nnNet~\cite{nonewnet}; (e) aeUnet \cite{brats_autoencoder}; (f) RandLA-Net~\cite{mrbrains18}; (g) our Point-Unet. Our method provides a good boundary on the tumor areas (pink arrows \textcolor{magenta}{$\rightarrow$}) compared with existing methods.}
\label{fig:visual}
\end{figure}

Figure~\ref{fig:visual} illustrates volumetric segmentation at three planes (sagittal, coronal, axial) on different methods given an input. The proposed Point-Unet (in the last column) provided a better segmentation results, resulting in better Dice score and Hausdorff95 distance. As can be seen, our Point-Unet segmentation provides an improved segmentation along the tumor boundary (indicated by the pink arrows) than the existing SOTA methods.

\bibliographystyle{splncs04}
\bibliography{paper2053}